\pdfoutput=1

\documentclass[11pt]{article}

\usepackage[final]{acl}

\usepackage{times}
\usepackage{latexsym}
\usepackage{booktabs}
\usepackage{multirow}
\usepackage{algorithm}
\usepackage{algorithm}
\usepackage{algpseudocode}
\usepackage[colorinlistoftodos]{todonotes}

\usepackage[T1]{fontenc}

\usepackage[utf8]{inputenc}

\usepackage{microtype}

\usepackage{inconsolata}

\usepackage{graphicx}
\usepackage{colortbl}
\definecolor{green2}{HTML}{BFD8B6}
\definecolor{green3}{HTML}{E7F0E5}
\definecolor{greenarrow}{HTML}{1DB100}
\definecolor{red3}{HTML}{C82506}

\usepackage{listings}
\usepackage{tcolorbox}
\definecolor{dkgreen}{rgb}{0,0.6,0}
\definecolor{gray}{rgb}{0.5,0.5,0.5}
\definecolor{mauve}{rgb}{0.58,0,0.82}
\lstdefinestyle{jinja2}{
  language={},
  keywordstyle={},
  basicstyle=\ttfamily\small,
  breaklines=true,
  breakatwhitespace=true,
  commentstyle={},
  stringstyle={}.
  identifierstyle={},
  showstringspaces=false,
  morekeywords={},
  literate=
    *{\{}{{\textcolor{red}{\{}}}{1}
     {\}}{{\textcolor{red}{\}}}}{1}
     {\_}{{\textcolor{blue}{\_}}}{1}
     {\%}{{\textcolor{red}{\%}}}{1},
}

\lstdefinelanguage{json}{
    basicstyle=\normalfont\ttfamily,
    commentstyle=\color{eclipseStrings}, 
    stringstyle=\color{eclipseKeywords}, 
    numbers=left,
    numberstyle=\scriptsize,
    stepnumber=1,
    numbersep=8pt,
    showstringspaces=false,
    breaklines=true,
    frame=lines,
    backgroundcolor=\color{gray}, 
    string=[s]{"}{"},
    comment=[l]{:\ "},
    morecomment=[l]{:"},
    literate=
        *{0}{{{\color{numb}0}}}{1}
         {1}{{{\color{numb}1}}}{1}
         {2}{{{\color{numb}2}}}{1}
         {3}{{{\color{numb}3}}}{1}
         {4}{{{\color{numb}4}}}{1}
         {5}{{{\color{numb}5}}}{1}
         {6}{{{\color{numb}6}}}{1}
         {7}{{{\color{numb}7}}}{1}
         {8}{{{\color{numb}8}}}{1}
         {9}{{{\color{numb}9}}}{1}
}

\tcbset{
    outer box/.style={
        enhanced,
        colback=white,
        colframe=black,
        boxrule=0.5pt,
        left=0pt,
        right=0pt,
        top=5pt,
        bottom=5pt,
        sharp corners,
        fonttitle=\bfseries\large,
        attach boxed title to top left={xshift=0.5cm, yshift=-\tcboxedtitleheight/2},
        boxed title style={sharp corners, colback=white}
    },
    inner box/.style={
        enhanced,
        colback=white,
        colframe=black!30,
        boxrule=0.5pt,
        left=10pt,
        right=10pt,
        top=10pt,
        bottom=10pt,
        arc=5pt,
        outer arc=5pt
    }
}

\title{VERITAS: A Unified Approach to Reliability Evaluation}

\author{Rajkumar Ramamurthy\thanks{Equal contribution} \quad Meghana Arakkal Rajeev$^\star$\quad Oliver Molenschot \\ \qquad \textbf{James Zou} \qquad \textbf{Nazneen Rajani}\\
\\
Collinear AI \\
\texttt{ \href{mailto:team@collinear.ai}{team@collinear.ai}} 
}

\begin{document}
\maketitle
\begin{abstract}

Large language models (LLMs) often fail to synthesize information from their context to generate an accurate response. This renders them unreliable in knowledge intensive settings where reliability of the output is key. A critical component for reliable LLMs is the integration of a robust fact-checking system that can detect hallucinations across various formats. While several open-access fact-checking models are available, their functionality is often limited to specific tasks, such as grounded question-answering or entailment verification, and they perform less effectively in conversational settings. On the other hand, closed-access models like GPT-4 and Claude offer greater flexibility across different contexts, including grounded dialogue verification, but are hindered by high costs and latency. In this work, we introduce VERITAS, a family of hallucination detection models designed to operate flexibly across diverse contexts while minimizing latency and costs. VERITAS achieves state-of-the-art results considering average performance on all major hallucination detection benchmarks, with $10\%$ increase in average performance when compared to similar-sized models and get close to the performance of GPT4 turbo with LLM-as-a-judge setting.

\end{abstract}

\section{Introduction}

Large language models (LLMs) \citep{brown2020languagemodelsfewshotlearners} have made remarkable strides in knowledge intensive tasks such as search, question answering, and natural language understanding. These models, trained on vast amounts of data, possess the ability to generate coherent and contextually relevant text. However, they also exhibit a concerning issue: their generated content often includes plausible but factually incorrect information. These incorrect outputs, known as \textit{hallucinations}, have raised increasing concerns about the safety and reliability of LLM applications \citep{xu2024hallucinationinevitableinnatelimitation}. We note that, although, hallucinations are undesired for knowledge intensive tasks, the same quirk is actually desirable in creative tasks such as story-telling, image-generation, prose or poetry generation, and brainstorming.

Hallucinations are particularly common in closed-book settings, where the knowledge encoded in the model’s weights and the LLM has to recall it during generation. In this setting, there is no relevant source material or context provided to the LLM and it has to solely rely on the knowledge embedded in its weight during the pre-training and post-training stages. ~\citet{kadavath2022languagemodelsmostlyknow} show that for such closed-book settings, self-evaluation, wherein a LLM evaluates the validity of its own claims and predicting whether it can correctly answer a user question, works well in the multiple-choice and true/false task settings.

On the other hand, in open-book settings, the LLM has access to relevant source materials either provided directly in context or as part of a compound system that uses Retrieval-Augmented Generation (RAG). RAG~\citep{guu2020realmretrievalaugmentedlanguagemodel, 10.5555/3495724.3496517, 10.5555/3648699.3648950} is a widely used approach for building user facing NLP applications, such as systems for grounded question answering (QA), fact-checking, summarization and customer
support. LLMs are often unreliable in such open-book settings, and generate responses that contradicts information provided in its context, especially when the knowledge domain of the information in context is out of distribution~\citep{shuster-etal-2021-retrieval-augmentation, magesh2024hallucinationfreeassessingreliabilityleading}. This severely hinders the application of these powerful models on private knowledge stores on complex data involving multi-step reasoning.

Many different reasons have been proposed for why LLMs exhibit such quirky behavior including defective training data and benchmarks~\citep{dziri-etal-2022-origin}, bias in training data~\citep{mckenna2023sources}, post-training on new knowledge~\citep{gekhman2024doesfinetuningllmsnew}, and knowledge cutoff~\citep{vu2023freshllmsrefreshinglargelanguage}.
To address the problem of reliability in LLMs, a number of benchmarks to evaluate factuality have been created in diverse task formats such as NLI, QA, and dialog. Humans or LLMs in the \textit{LLM-as-a-Judge} settings are primarily used for evaluation on these benchmarks. It leverages the power of LLMs to perform the role of \textit{Judges}, which can provide judgements on content quality, coherence, and alignment~\citep{vu2024foundationalautoraterstaminglarge,kim2024prometheusinducingfinegrainedevaluation}. Lynx~\citep{ravi2024lynxopensourcehallucination} and  BeSpoke-MiniCheck~\citep{tang2024minicheck} are examples of LLM-as-a-judge trained to judge hallucinations for QA and NLI tasks respectively. These models do no generalize to tasks beyond what they are trained on.
Consequently, there is no single model that works across different task formats and on diverse domain datasets.

A unified approach to evaluating LLM reliability would not only enable better understanding of gaps in capabilities of LLMs, but also to develop more robust post-training techniques including RLAIF (reinforcement learning with AI feedback)~\citep{lee2024rlaif} wherein factual responses are chosen over non-factual ones. As a result, the ability of the AI judge to distinguish between factual and non-factual responses by cross-checking against retrieved documents is critical for this alignment process \citep{tian2023finetuninglanguagemodelsfactuality, lin2024flamefactualityawarealignmentlarge}.
\noindent
Our \textbf{contributions} are three-fold:
\begin{itemize}
  \item We unify the hallucination detection problem and propose a multi-task setup that includes NLI, QA, and dialog. Our VERITAS judge is state-of-the-art on LLM-AggreFact, Halubench, HalluDial, and two proprietary enterprise datasets, reaching the performance of GPT4 Turbo.
  \item We curate VERITAS Collection, high quality, diverse training dataset covering different formats for training hallucination detection models.
  \item We present VERITAS Bench, a unified benchmark for evaluating LLMs on reliability in open-book settings across 18 different datasets covering three different tasks.
\end{itemize}

\begin{figure*}[t]
    \centering
    \includegraphics[width=0.93\linewidth]{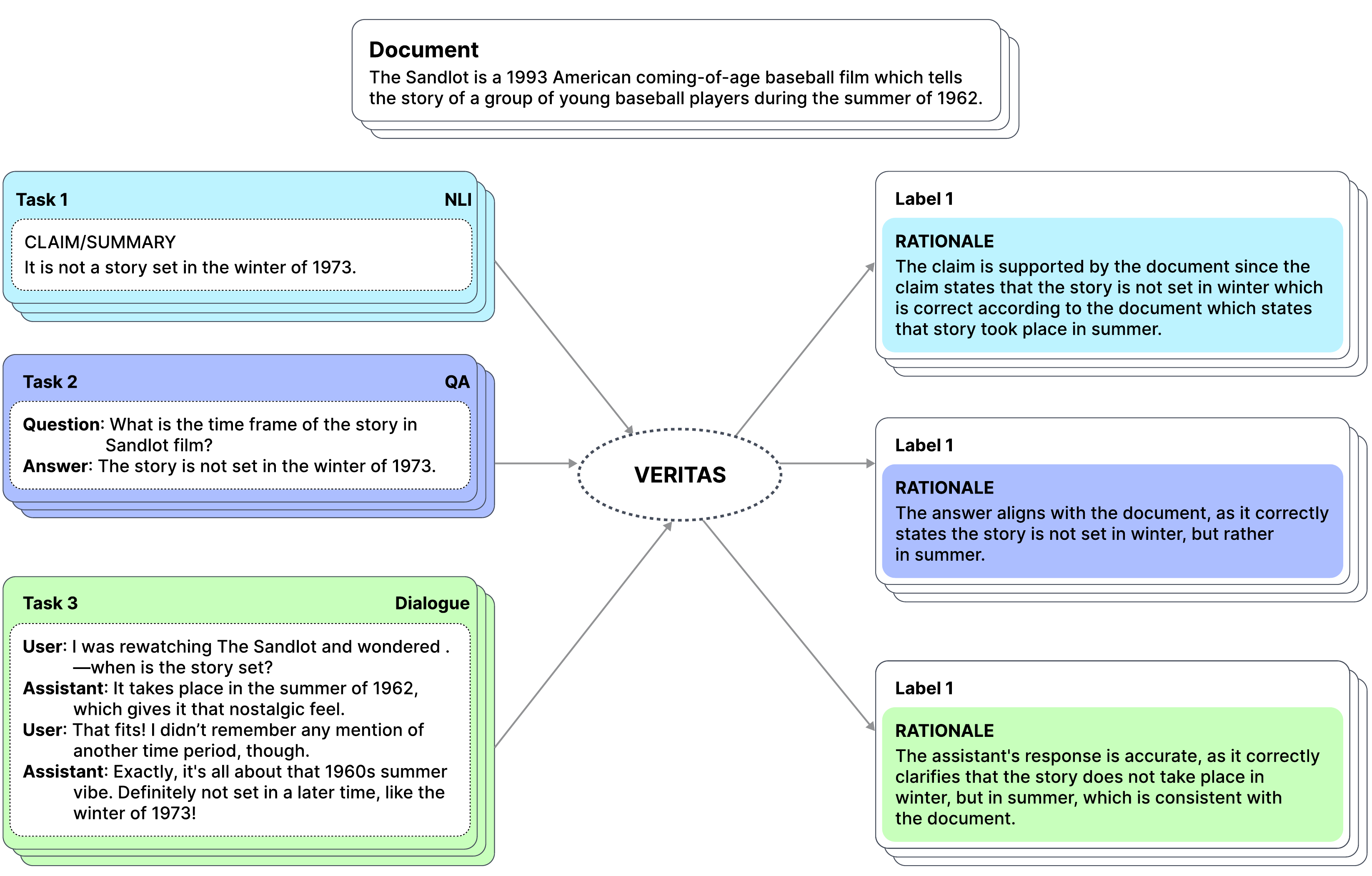}
    \caption{VERITAS models provide a unified interface for hallucination detection using a multi-task approach comprising of three tasks. 1) \textbf{NLI task} in which the claim or the summary of the given document is checked/ verified 2) \textbf{Grounded QA} in which the answer is assessed for factuality 3) \textbf{Grounded Dialogue} in which the assistant responses are verified. In all these tasks, evaluation is performed based on the given document. Label 1 indicates no hallucination (content is factually consistent), while Label 0 denotes factual inconsistencies.}
    \label{fig:unified_format}
\end{figure*}

\section{Related Work}
\paragraph{\textbf{Reliability Evaluation}}
There has been a slew of work on LLM evaluation and there is an entire line of research that focuses on reliability or hallucination evaluation. 
The hallucination evaluation task is to predict whether or not the LLMs response is consistent with a given context. The context is usually in the form of a long-form source document or $k$ retrieved documents from a RAG retriever. The task setup for this kind of evaluation is like the \textit{open-book exam} wherein the LLM gains new knowledge through the context. In contrast, the \textit{closed-book exam} evaluates the ability of a LLM to retrieve knowledge accurately from its weights baked-in during pre-training or post-training. 

The hallucination evaluators belong to two broad classes of models: 1. classifier models (encoder transformers) and 2. generative models (decoder transformers).

ARES~\citep{saadfalcon2024aresautomatedevaluationframework} is a suite of classifier models based on DeBERTa-v3-
Large finetuned with synthetic data constructed by bootstrapping few-shot demonstrations. ~\citet{belyi2024lunaevaluationfoundationmodel} proposed a similar DeBERTa-based proprietary classifier model, Luna, finetuned on proprietary datasets across different industry segments. Minicheck-DBTA is also a DeBERTA--v3-Large finetuned for fact-checking by training on a combination of synthetic and a subset of adversarial NLI dataset~\citep{tang2024minicheckefficientfactcheckingllms}.

Generative hallucination evaluators include the LLM-as-a-Judge models, that are not specialized for reliability, such as JudgeLM~\citep{zhu2023judgelmfinetunedlargelanguage}, Prometheus~\citep{kim2024prometheusinducingfinegrainedevaluation}, and others~\citep{zheng2023judgingllmasajudgemtbenchchatbot,gao2023rarrresearchingrevisinglanguage}. Lynx~\citep{ravi2024lynxopensourcehallucination} is a hallucination detector LLM based on the Llama-3-Instruct models in 8B and 70B, finetuned on $2400$ training examples sampled from FinanceBench~\citep{islam2023financebenchnewbenchmarkfinancial}, DROP~\citep{dua2019dropreadingcomprehensionbenchmark}, CovidQA~\citep{moller-etal-2020-covid} and PubMedQA~\citep{jin2019pubmedqadatasetbiomedicalresearch} wherein half of those examples are perturbed to construct hallucinated answers.

Apart from the above two categories of evaluators, yet another line of work focuses on heuristic-based consistency checks~\citep{agrawal-etal-2024-language, manakul2023selfcheckgptzeroresourceblackboxhallucination, guerreiro2023lookingneedlehaystackcomprehensive}.
\paragraph{\textbf{Benchmarks and Metrics}}
HaluEval~\citep{li-etal-2023-halueval} is a collection of 35k samples consisting of QA and dialog tasks created using a two-step process of filtering and annotations for hallucinations. On the other hand, HaluBench~\citep{ravi2024lynxopensourcehallucination} is a a QA hallucination evaluation benchmark of 15k samples that consists of context along with a question-answer pair and whether the answer contains hallucination or not.
HalluDial~\citep{luo2024halludiallargescalebenchmarkautomatic} is a benchmark focused on dialog hallucination detection with about 4k dialogs with an average of 4.5 turns. The benchmark comprises of conversations across seven categories: game, food, music, culture, health, animal, and color.
LLM-AggreFact~\citep{tang2024minicheckefficientfactcheckingllms} is a benchmark consisting of ten publicly available datasets annotated by humans for fact checking consistency.

EXAM is a metric that estimates number of questions answered correctly, for a set of queries, by a QA system based on the LLM response ~\citep{exam}.
ARES proposes a score for RAG system ranking based on its judge models using prediction-powered inference (PPI)~\citep{angelopoulos2023predictionpoweredinference} that leverages the human annotated samples for computing confidence intervals~\citep{saadfalcon2024aresautomatedevaluationframework}.
RAGAS~\citep{es2023ragasautomatedevaluationretrieval} used LLMs to generate statements form a question-answer pair computes three evaluation metrics, faithfulness, answer relevance, and context relevance. Other benchmarks for evaluating knowledge synthesis and factuality in LLMs include the KILT~\citep{petroni-etal-2021-kilt}, TruthfulQA~\citep{lin-etal-2022-truthfulqa}, BEGIN~\citep{10.1162/tacl_a_00506} and FreshQA~\citep{vu2023freshllmsrefreshinglargelanguage}. 

\section{VERITAS Data Collection}

Addressing the limitations of current open source models, our primary goal is to design a model that is capable of flexibly handling various input formats. To achieve this, we aim to curate a diverse training data that comprises of tasks like textual entailment, summarization, question answering and grounded dialogue verification. 

\subsection{Textual Entailment} For textual entailment, we primarily source data from the ANLI dataset \citep{nie-etal-2020-adversarial}, which consists of adversarially constructed entailment data points. Each data point includes a premise, a hypothesis, and a label indicating one of three relationships: entailment, contradiction, or neutral. To focus on more informative signals, we use data from the second iteration of ANLI, excluding neutral instances to emphasize clearer learning patterns. Additionally, we incorporate existing datasets for hallucination detection, particularly Minicheck \citep{tang2024minicheck}, which provides document-claim pairs with a focus on sentence-level claims. Both datasets are reformatted into a unified structure of (document, claim, label) to ensure consistency.

\subsection{Grounded Question Answering} We repurpose existing reading comprehension datasets consisting of passage, question, and answer tuples, each emphasizing different aspects of evaluation.

\begin{itemize}
    \item \textbf{DROP} \citep{dua2019dropreadingcomprehensionbenchmark}: This dataset contains passages, questions, and answers, where verifying the correctness of an answer often requires complex numerical reasoning. The task involves extracting numerical facts from various parts of the passage and performing operations like addition, counting, and sorting.
    \item \textbf{NewsQA}~\citep{trischler2016newsqa}: Built on CNN news articles, this machine comprehension dataset provides question-answer pairs. It poses challenges as it requires reasoning beyond simple word matching or entailment.
    \item \textbf{TriviaQA}~\citep{joshi2017triviaqa}: This dataset consists of question-answer pairs grounded in Wikipedia and web articles written by trivia enthusiasts. We specifically consider samples grounded in web articles, as their unstructured nature helps in detecting hallucinations in real-world settings.
    \item \textbf{SearchQA}~\citep{dunn2017searchqa}: Designed to simulate an end-to-end QA system, this dataset consists of question-answer pairs grounded in noisy documents, which may include irrelevant information. Similar to TriviaQA, this setup closely resembles a typical retrieval-augmented generation (RAG) system.
\end{itemize}

Additionally, we curate a validation split consisting of samples from DROP, TextBookQA~\citep{Kembhavi2017AreYS}, and RACE~\citep{lai2017race}, ensuring evaluation on out of domain datasets.

\subsection{Summarization}

For the summarization component, we curate a diverse collection of datasets that span multiple domains and styles. The datasets are selected to represent different types of source documents. \\
The different datasets used are: BillSum \citep{DBLP:journals/corr/abs-1910-00523} containing US congressional and California state bills and their summaries; SAMSum \citep{DBLP:journals/corr/abs-1911-12237} comprising messenger-like conversations with abstractive summaries; BigPatent \citep{DBLP:journals/corr/abs-1906-03741} with 1.3 million U.S. patent documents and their human-written summaries; and Multi-News \citep{DBLP:journals/corr/abs-1906-01749}, which features human written summaries of news articles from newser.com. \\
We sample equally from these datasets. forming ~12k datapoints to form our summarization task. Additionally, we augment the summarization dataset with negative examples by generating factually unsupported summaries, as all original summaries in these datasets are inherently factually supported. \\ \\
Additionally, to bring the same domain diversity to the NLI task and to maximize the utility of this data, we also reformulate the summarization instances as NLI tasks by treating summaries as claims, thereby enriching our NLI task.


\begin{table}[t!]
\centering
\resizebox{0.8\columnwidth}{!}{%
\begin{tabular}{l|cc|cc}
\toprule
\textbf{Format} & \multicolumn{2}{c}{\textbf{Train}} & \multicolumn{2}{c}{\textbf{Dev}} \\
\cmidrule(lr){2-3} \cmidrule(lr){4-5} \\ 
& En & Esp & En & Esp \\
\midrule
NLI & 50967 & 30897 & 3986 & 542\\
Summarization & 17056 & - & 3416& -\\
QA & 26433 & 3807 & 1739 & 596\\
Dialogue & 25878 & 2432 & 1666 & 279\\
\midrule
\textbf{Total}            & 120334           & 37136 & 10807 & 1417 \\ 
\bottomrule
\end{tabular}%
}
\caption{\textbf{VERITAS Training collection}: Overview of curated dataset with counts of different splits, and sub-splits.} \label{tab:dataset_overview}
\end{table}

\subsubsection{Generation of hallucination samples} Since the aforementioned datasets are primarily question-answer datasets, they do not naturally include answers that are unfaithful or incorrect relative to the context. To address this, we generate incorrect answers using Llama 70B Instruct, prompting it to produce unfaithful answers based on the given context. 

We generate diverse hallucination types following the taxonomy proposed by \citet{mishra2024finegrainedhallucinationdetectionediting}, which includes: \textit{Entity errors}, where an incorrect entity alters the factuality of a statement; \textit{Relation errors}, involving incorrect semantic relationships like verbs or prepositions; \textit{Sentence errors}, where the entire statement contradicts the evidence; \textit{Invented errors}, containing fabricated information not found in the context; \textit{Subjective errors}, based on personal opinions rather than facts; and \textit{Unverifiable errors}, where the answer cannot be validated by the given evidence. This variety ensures we generate diverse factual errors for effective hallucination detection. 

\subsection{Grounded Dialogue Verification}

To train models that can detect factual errors in conversational settings, we aim to include grounded dialogue verification task. In this task, there is a conversation between user and assistant, where the responses by assistant are evaluated against a reference document (e.g., RAG context). Since no publicly available datasets contain document-conversation-label triples for this purpose, we generate this dataset synthetically. To achieve this, we convert the curated QA samples into dialogue by prompting Llama 70B Instruct to transform question-answer pairs into multi-turn conversations between a user and an assistant, ensuring that the factual information in the answers is preserved. 

\subsection{Spanish data}
We translated data from English to Spanish to improve the judge's proficiency in Spanish. This translation process included datapoints in NLI, QA, and Dialogue formats, with 37k for training and 1.5k for validation. The LLama3.1-8B Instruct model was used for translating these datapoints.


\subsection{Rationale Generation and Data Cleansing}

In order to train models that can learn from reasoning paths, we generate rationales for the collected dataset. These rationales provide a clear explanation of the logic behind each response, allowing the model to understand not only the correct label but also the reasoning that supports it. Beyond simple explanations, rationales act as critical learning signals for the models. To generate these rationales, we prompt GPT-4o to output both the rationale and the label for each data point multiple times. If the output labels consistently contradict the ground truth label, we discard the data point due to this inconsistency. When the labels align with the ground truth,  we retain the rationale which corresponding to the correct prediction. This approach helps to ensure high-quality rationales and accurate labels. 

\section{VERITAS}

We fine-tune two classes of transformer models by leveraging VERITAS Data: encoder-based classifier and decoder-based generative models. 

\subsection{Classifier models} For the classifier model, we fine-tune DeBERT-v3-large~\citep{he2023debertav3improvingdebertausing} as this backbone has demonstrated strong performance in claim verification tasks in previous works \citep{tang2024minicheck, belyi2024lunaevaluationfoundationmodel}. To unify the input text format, we convert each data point into (document, conversation) structure applicable for all formats including NLI, QA and dialogue. For QA, the claim is formatted as a single-turn dialogue between the user and the assistant. In the case of NLI, the claim is presented as the assistant response. Please refer to Appendix \ref{sec:appendix_deberta_input_format} for exact input formatting. We use the standard cross-entropy loss for training and fine-tune the model for $2$ epochs with a constant learning rate of $1e^{-6}$, a warm up ratio $0.1$, batch size of $4$, weight decay of $0.001$. Hereafter, we refer to this resulting classifier as VERITAS DeBERTa.

\subsection{Generative models}
\label{sec:training_gen_models}

Next, we considered generative transformer models, as they are increasingly applied to tasks such as LLM-as-judges, including hallucination detection. For this, we fine-tune variants of LLaMA 3.2 3B and LLaMA 3.1 8B Instruct models. Given the strong instruction-following capabilities of these models, we adopt a multi-task setup without the need for strict unified formatting. Instead, we use task-specific instruction templates, tailored for each task while maintaining consistency across the templates by varying only the entity being assessed. This flexibility ensures that the models can handle different types of inputs effectively without compromising on performance. For detailed input formatting and instruction templates, please refer to Appendix \ref{sec:appendix_llama_input_format}. 


We fine-tune these models using teacher forcing, where the models generate JSON-formatted outputs that include both the rationale and the label for the entire training split of VERITAS Data, whose train and dev compositions are shown in Table \ref{tab:dataset_overview}. Each example is structured in a chat-based format, with the input appearing as a user query and the corresponding ground truth label and rationale in the assistant's response. To reduce computational costs, we adopt QLoRA~\citep{dettmers2024qlora}, fine-tuning the models in a 4-bit format. We set the low-rank adapter rank to 64, specifically targeting key transformer components such as projection layers, embeddings, and the final language model layers. Training is performed with an 8192-token sequence length and a learning rate of $5e^{-6}$, along with a warm-up ratio of 0.1, for a single epoch. We apply gradient accumulation over two steps with a batch size of 2. 
The resulting models are referred to as VERITAS 3B and VERITAS 8B.


\begin{table}[t!]
\centering
\resizebox{0.8\columnwidth}{!}{%
\begin{tabular}{l|l|c}
\toprule
\textbf{Format} & \textbf{Dataset} & \textbf{Count}\\ \toprule
\multirow{1}{*}{NLI}      & LLMAggreFact  & 29320\\ 
\midrule
\multirow{4}{*}{QA}       & Halu Bench PubMedQA    & 1000 \\
                          & Halu Bench FinanceBench & 1000 \\ 
                          & HaluEval QA    & 20000 \\ 
\midrule
\multirow{2}{*}{Dialogue}  & HaluEval Dialog  & 20000 \\
                          & HalluDial  & 10000 \\ 
\midrule
\multirow{2}{*}{Enterprise}  & Real Estate QA   & 108 \\
                          & Persona Chatbot Dialog  & 92 \\ 
\midrule
\textbf{Total}            & -           & 81520 \\ 
\bottomrule
\end{tabular}
}
\caption{\textbf{VERITAS Bench} Comprehensive benchmark for hallucination detection consisting of three different evaluation formats}
\label{tab:veritasbench_overview}
\end{table}

\begin{table*}[t!]
\centering
\scriptsize
\resizebox{0.98\textwidth}{!}{
\begin{tabular}{l|cccccc|c}
\textbf{Model}&\multicolumn{1}{c}{\textbf{NLI}}&\multicolumn{3}{c}{\textbf{QA}}&\multicolumn{2}{c}{\textbf{Dialog}}&\textbf{Average}\\
\cmidrule(lr){2-2} \cmidrule(lr){3-5} \cmidrule(lr){6-7} 
&LLM AggreFact&PubMedQA&FinanceBench&HaluEvalQA&HalluDial&HaluEval Dialog\\
\toprule
GPT4 Turbo&76.1&86.8&73.9&85.6&76.0&69.2&77.9\\
\midrule
\textit{\underline{Classifiers}}&&&&&&\\ 
Minicheck DeBERTa (440M)&73.1&59.7&50.0&63.8&52.0&50.5&58.2\\
VERITAS DeBERTa (440M) &73.0&75.5&56.7&84.6&73.2&65.9&\textbf{71.5}\\ 
\midrule
\textit{\underline{Generative}}&&&&&&\\
Lynx 8B&69.3&88.6&75.4&87.2&66.7&68.2&75.9\\
Bespoke MiniCheck 7B&77.4&75.2&58.1&87.2&64.6&55.3&69.6\\
\cmidrule(lr){2-8}
VERITAS 3B&73.2&79.9&65.0&83.7&74.0&67.7&73.9\\
VERITAS 8B &74.0&83.9&68.2& 84.9&73.6&74.5&\textbf{76.5}\\

\bottomrule
\end{tabular}
}

\caption{\textbf{VERITAS Bench Results}: Main results comparing VERITAS models to existing open and closed-access models across the three task formats, NLI, QA, and dialog. While Lynx, Bespoke MiniCheck 7B excel at QA and NLI tasks respectively, they do not generalize on conversational settings. On the other hand, VERITAS models perform well on all formats. 
}
\label{table:main}
\end{table*}


\section{VERITAS Bench}

\begin{table*}[t!]
\centering
\resizebox{0.99\textwidth}{!}{
\begin{tabular}{l|ccccccccccc|c}
\textbf{Model}&\multicolumn{12}{c}{\textbf{LLMAggreFact}}\\
\cmidrule(lr){2-13}
&AggreFact-CNN&AggreFact-XSum&TofuEval-MediaS&TofuEval-MeetB&Wice&Reveal&ClaimVerify&FactCheck-GPT&ExpertQA&Lfqa&RAGTruth&Average\\ \\
\toprule
GPT4 Turbo&65.3&73.7&72.4&82.0&76.8&88.7&69.3&80.0&60.6&83.3&85.5&76.1\\
\midrule
\textit{\underline{Classifiers}}&&&&&&\\ 
Minicheck DeBERTa&64.2&71.0&69.3&72.7&69.4&87.3&75.6&73.0&58.9&83.9&78.8&73.1\\
VERITAS DeBERTa& 54.6&73.5&\textbf{72.7}&77.8&74.8&84.2&70.7&74.3&59.2&85.9&75.5&73.0\\
\midrule
\textit{\underline{Generative}}&&&&&&\\
Lynx 8B&59.3&64.7&70.7&77.6&65.4&74.7&69.7&67.9&58.6&77.9&76.1&69.3\\
Bespoke MiniCheck 7B&65.5&77.8&76.0&78.3&83.0&88.0&75.3&77.7&59.2&86.7&84.0&77.4\\
\cmidrule(lr){2-13}
VERITAS 3B&52.3&71.7&72.3&74.8&72.5&88.1&74.7&78.3&59.4&86.5&75.0&73.2\\
VERITAS 8B&53.9&71.8&74.0&78.3&74.9&87.8&71.5&79.3&60.0&83.7&79.3&74.0 \\
\\

\bottomrule
\end{tabular}
}
\caption{\textbf{LLMAggreFact}: Results comparing VERITAS models to existing open and closed-access models on subsplits of LLMAggreFact}
\label{table:llm_aggrefact}
\end{table*}

Existing hallucination benchmarks have primarily focused on specific types of evaluation. For example, LLMAggreFact \citep{tang2024minicheck} centers exclusively on claim verification, while benchmarks like HaluBench~\citep{ravi2024lynxopensourcehallucination} target question-answering (QA) tasks. Although some benchmarks for grounded dialogue verification exist~\citep{luo2024halludiallargescalebenchmarkautomatic}, they are rarely utilized in the development of fact-checking models. This gap motivated us to design a comprehensive benchmark that incorporates three evaluation formats: claim verification, question answering, and dialogue verification, providing a more holistic assessment of model performance. The exact composition of different tasks in VERITAS Bench is depicted in Table \ref{tab:veritasbench_overview}.

\paragraph{Claim Verification} For the claim verification task, we leverage LLMAggreFact \citep{tang2024minicheck}, as it contains a wide array of subtasks within. It provides a diverse set of claims that cover multiple scenarios, including claims derived from model-generated summaries, LLM responses to search queries and claims sourced from Wikipedia. This diversity ensures a comprehensive evaluation of models for claim verification across various contexts and content types.

\paragraph{Question Answering} For the question-answering format of evaluation, we include two key benchmarks: HaluEval~\citep{li-etal-2023-halueval} and HaluBench~\citep{ravi2024lynxopensourcehallucination}. In HaluEval, we focus solely on the QA-type samples. Similarly for HaluBench, we incorporate datasets such as PubMedQA~\citep{jin2019pubmedqadatasetbiomedicalresearch}, FinanceBench~\citep{islam2023financebenchnewbenchmarkfinancial} 
, which provide domain-specific challenges for QA. Notably, we exclude any samples from the DROP dataset, as some of VERITAS training data is derived from it. Additionally, we exclude splits of RAGTruth ~\citep{wu2023ragtruth} and HaluEval are excluded as they are already included in under other splits.

\paragraph{Grounded Dialogue Verification} We consider the dialogue split from HaluEval and include HalluDial~\citep{luo2024halludiallargescalebenchmarkautomatic}, one of the most extensive benchmarks for hallucination detection in dialogue contexts. It is to be noted that test split of HalluDial is not publicly available, so we use random subset containing 10k samples from train split. 

\paragraph{Enterprise} Additionally, we include two proprietary enterprise datasets from domains of persona chatbots and real estate QA system covering dialogue and QA formats respectively.

\section{Results}
We benchmarked the VERITAS models on the VERITAS Bench by comparing them against several baselines, including current state-of-the-art classifier and generative models. In the generative models category, we selected GPT-4 Turbo as the closed-source baseline due to its competitive performance across various LLM-as-judge benchmarks. Additionally, we included Lynx 8B~\citep{ravi2024lynxopensourcehallucination} and Bespoke MiniCheck 7B~\citep{tang2024minicheck}, both of which have shown impressive results on QA and NLI evaluations respectively. For classifier models, we primarily considered Minicheck DeBERTa, known for its strong performance in claim verification tasks. As evaluation metrics, we use balanced accuracy for LLMAggreFact and for all others, we report accuracy as the main evaluation metric.

\subsection{Main Results}

\begin{table*}[t!]
\centering
\small
\begin{tabular}{l|ccc|c}
\textbf{Model} & \multicolumn{2}{c}{\textbf{Real Estate} (QA)} & \textbf{Persona Chatbots} (Dialog) & \textbf{Average} \\
\cmidrule(lr){2-3}\cmidrule(lr){4-4}\cmidrule(lr){5-5}\\
& English & Spanish & & \\
\toprule
GPT4 Turbo & 66.1 & 56.1 & 90.2 & 70.8 \\
\midrule
\textit{\underline{Baselines}}\\ 
Lynx 8B & 64.7 & 55.6 & 80.4 & 66.9 \\
Bespoke MiniCheck 7B & 65.7 & 50.5 & 55.4 & 57.2 \\
Minicheck DeBERTa & 73.8 & 57.0 & 62.0 & 64.3 \\
\cmidrule(lr){2-5}
VERITAS DeBERTa & 77.4 & 70.8 & 91.3 & 79.8 \\
\midrule
Veritas 3B & 72.7 & 73.7 & 90.2 & 78.9 \\
Veritas 8B & 83.2 & 72.2 & 90.2 & \textbf{81.9} \\
\bottomrule
\end{tabular}
\caption{Results comparing Veritas models to existing open and closed-access models on Enterprise proprietary datasets in the domains of persona chatbots and real estate}
\label{table:enterprise}
\end{table*}

\paragraph{VERITAS models generalize across all formats} Table~\ref{table:main} presents the key results from VERITAS Bench. Minicheck DeBERTa performs well in NLI but falls behind in QA and dialogue tasks, primarily because it is exclusively trained on claim verification, making it less effective for handling QA and conversation-based inputs. In contrast, VERITAS DeBERTa, trained using a multi-task approach, surpasses Minicheck in both QA and dialogue formats. Among generative models, Lynx 8B and Bespoke MiniCheck 7B excel in QA but struggle with dialogue tasks. Lynx's strong QA performance can be attributed to its training data, which is partially sourced from PubMedQA. Similarly, training data of Bespoke MiniCheck 7B largely involves NLI samples, alongside proprietary data whose format is not publicly known. Our VERITAS 8B, trained on a more diverse dataset, performs competitively, nearly closing the gap with GPT-4.

\paragraph{Encoder models are natural fact checkers} DeBERTa-based models, despite having relatively fewer parameters (440M), perform exceptionally well on benchmarks such as LLMAggreFact (see Table~\ref{table:llm_aggrefact}). Interestingly, generative models such as LLama 3B, even when trained with reasoning traces, could not match the performance of VERITAS DeBERTa. There could be several reasons for this. First, encoder models excel in entailment based tasks, making them naturally well-suited for claim verification. Second, training generative models to predict a final label is inherently challenging. While rationales are helpful, the actual label appears at the end, which might lead to focus more on mimicking the style of rationales instead of learning the correct output. A potential solution to this issue could be augmenting the data with examples that omit rationales~\citep{wang2024direct}, allowing the model to better learn how to produce accurate labels.


\paragraph{On enterprise data} 

As show in Table~\ref{table:enterprise}, the VERITAS models, particularly VERITAS 8B, demonstrate exceptional performance across both real estate QA and persona chatbot dialogue tasks. Similarly VERITAS 3B also performs strongly surpassing much larger models such as Lynx 8B and Minicheck 7B. Both models surpass all baselines and notably VERITAS 8B outperforms even GPT-4 Turbo highlighting its effectiveness in handling multi-lingual and diverse formats, showcasing its adaptability for entreprise-level applications.

\section{Conclusion}
VERITAS is a unified approach for judging reliability of LLMs that employs a multi-task training setup across NLI, QA, and dialog. Our approach outperforms existing open-access models while maintaining competitive performance to GPT4 turbo with blazing fast inference ($\sim$ 100 milliseconds latency) and low costs. The generative judges can learn and self-improve from rationale without any additional training data. 

VERITAS bench is a unified benchmark for evaluating hallucination across 18 different datasets from different domains. Our results on the VERITAS benchmark confirm that encoder models such as DeBERTa are natural fact checkers and perform competitively with models that are 16x their size. The VERITAS collection and benchmark lay the groundwork for robust post-training techniques of LLMs that rely on AI feedback for aligning LLMs to make them more truthful. 

\section*{Limitations}

\paragraph{Document Length} One key limitation of our approach is the document length constraint in generative models. Despite training with a sequence length of $8192$ tokens, handling much longer documents remain a challenge. This may limit performance on tasks requiring full-length articles or extensive reports, a few of them in LLMAggreFact benchmark. Currently, we mitigate this by splitting documents into smaller chunks and aggregating the results, following prior approaches~\citep{tang2024minicheck}. However, this solution is suboptimal.


\paragraph{Backbones} The backbone architectures used in VERITAS, such as LLama and DeBERTa, may not be fully optimized for the specialized task of reliability assessment. Exploring alternative architectures, like Flan-T5, could lead to performance gains~\citep{tang2024minicheck}, especially in our multi-task setups. The multi-task instruction fine-tuning of Flan-T5~\citep{chung2024scaling} makes it a promising candidate for future work.

\bibliography{custom}

\begin{thebibliography}{56}
\providecommand{\natexlab}[1]{#1}

\bibitem[{Agrawal et~al.(2024)Agrawal, Suzgun, Mackey, and Kalai}]{agrawal-etal-2024-language}
Ayush Agrawal, Mirac Suzgun, Lester Mackey, and Adam Kalai. 2024.
\newblock \href {https://aclanthology.org/2024.findings-eacl.62} {Do language models know when they{'}re hallucinating references?}
\newblock In \emph{Findings of the Association for Computational Linguistics: EACL 2024}, pages 912--928, St. Julian{'}s, Malta. Association for Computational Linguistics.

\bibitem[{Angelopoulos et~al.(2023)Angelopoulos, Bates, Fannjiang, Jordan, and Zrnic}]{angelopoulos2023predictionpoweredinference}
Anastasios~N. Angelopoulos, Stephen Bates, Clara Fannjiang, Michael~I. Jordan, and Tijana Zrnic. 2023.
\newblock \href {https://arxiv.org/abs/2301.09633} {Prediction-powered inference}.
\newblock \emph{Preprint}, arXiv:2301.09633.

\bibitem[{Belyi et~al.(2024)Belyi, Friel, Shao, and Sanyal}]{belyi2024lunaevaluationfoundationmodel}
Masha Belyi, Robert Friel, Shuai Shao, and Atindriyo Sanyal. 2024.
\newblock \href {https://arxiv.org/abs/2406.00975} {Luna: An evaluation foundation model to catch language model hallucinations with high accuracy and low cost}.
\newblock \emph{Preprint}, arXiv:2406.00975.

\bibitem[{Brown et~al.(2020)Brown, Mann, Ryder, Subbiah, Kaplan, Dhariwal, Neelakantan, Shyam, Sastry, Askell, Agarwal, Herbert-Voss, Krueger, Henighan, Child, Ramesh, Ziegler, Wu, Winter, Hesse, Chen, Sigler, Litwin, Gray, Chess, Clark, Berner, McCandlish, Radford, Sutskever, and Amodei}]{brown2020languagemodelsfewshotlearners}
Tom~B. Brown, Benjamin Mann, Nick Ryder, Melanie Subbiah, Jared Kaplan, Prafulla Dhariwal, Arvind Neelakantan, Pranav Shyam, Girish Sastry, Amanda Askell, Sandhini Agarwal, Ariel Herbert-Voss, Gretchen Krueger, Tom Henighan, Rewon Child, Aditya Ramesh, Daniel~M. Ziegler, Jeffrey Wu, Clemens Winter, Christopher Hesse, Mark Chen, Eric Sigler, Mateusz Litwin, Scott Gray, Benjamin Chess, Jack Clark, Christopher Berner, Sam McCandlish, Alec Radford, Ilya Sutskever, and Dario Amodei. 2020.
\newblock \href {https://arxiv.org/abs/2005.14165} {Language models are few-shot learners}.
\newblock \emph{Preprint}, arXiv:2005.14165.

\bibitem[{Chung et~al.(2024)Chung, Hou, Longpre, Zoph, Tay, Fedus, Li, Wang, Dehghani, Brahma et~al.}]{chung2024scaling}
Hyung~Won Chung, Le~Hou, Shayne Longpre, Barret Zoph, Yi~Tay, William Fedus, Yunxuan Li, Xuezhi Wang, Mostafa Dehghani, Siddhartha Brahma, et~al. 2024.
\newblock Scaling instruction-finetuned language models.
\newblock \emph{Journal of Machine Learning Research}, 25(70):1--53.

\bibitem[{Dettmers et~al.(2024)Dettmers, Pagnoni, Holtzman, and Zettlemoyer}]{dettmers2024qlora}
Tim Dettmers, Artidoro Pagnoni, Ari Holtzman, and Luke Zettlemoyer. 2024.
\newblock Qlora: Efficient finetuning of quantized llms.
\newblock \emph{Advances in Neural Information Processing Systems}, 36.

\bibitem[{Dua et~al.(2019)Dua, Wang, Dasigi, Stanovsky, Singh, and Gardner}]{dua2019dropreadingcomprehensionbenchmark}
Dheeru Dua, Yizhong Wang, Pradeep Dasigi, Gabriel Stanovsky, Sameer Singh, and Matt Gardner. 2019.
\newblock \href {https://arxiv.org/abs/1903.00161} {Drop: A reading comprehension benchmark requiring discrete reasoning over paragraphs}.
\newblock \emph{Preprint}, arXiv:1903.00161.

\bibitem[{Dunn et~al.(2017)Dunn, Sagun, Higgins, Guney, Cirik, and Cho}]{dunn2017searchqa}
Matthew Dunn, Levent Sagun, Mike Higgins, V~Ugur Guney, Volkan Cirik, and Kyunghyun Cho. 2017.
\newblock Searchqa: A new q\&a dataset augmented with context from a search engine.
\newblock \emph{arXiv preprint arXiv:1704.05179}.

\bibitem[{Dziri et~al.(2022{\natexlab{a}})Dziri, Milton, Yu, Zaiane, and Reddy}]{dziri-etal-2022-origin}
Nouha Dziri, Sivan Milton, Mo~Yu, Osmar Zaiane, and Siva Reddy. 2022{\natexlab{a}}.
\newblock \href {https://doi.org/10.18653/v1/2022.naacl-main.387} {On the origin of hallucinations in conversational models: Is it the datasets or the models?}
\newblock In \emph{Proceedings of the 2022 Conference of the North American Chapter of the Association for Computational Linguistics: Human Language Technologies}, pages 5271--5285, Seattle, United States. Association for Computational Linguistics.

\bibitem[{Dziri et~al.(2022{\natexlab{b}})Dziri, Rashkin, Linzen, and Reitter}]{10.1162/tacl_a_00506}
Nouha Dziri, Hannah Rashkin, Tal Linzen, and David Reitter. 2022{\natexlab{b}}.
\newblock \href {https://doi.org/10.1162/tacl_a_00506} {{Evaluating Attribution in Dialogue Systems: The BEGIN Benchmark}}.
\newblock \emph{Transactions of the Association for Computational Linguistics}, 10:1066--1083.

\bibitem[{Es et~al.(2023)Es, James, Espinosa-Anke, and Schockaert}]{es2023ragasautomatedevaluationretrieval}
Shahul Es, Jithin James, Luis Espinosa-Anke, and Steven Schockaert. 2023.
\newblock \href {https://arxiv.org/abs/2309.15217} {Ragas: Automated evaluation of retrieval augmented generation}.
\newblock \emph{Preprint}, arXiv:2309.15217.

\bibitem[{Fabbri et~al.(2019)Fabbri, Li, She, Li, and Radev}]{DBLP:journals/corr/abs-1906-01749}
Alexander~R. Fabbri, Irene Li, Tianwei She, Suyi Li, and Dragomir~R. Radev. 2019.
\newblock \href {https://arxiv.org/abs/1906.01749} {Multi-news: a large-scale multi-document summarization dataset and abstractive hierarchical model}.
\newblock \emph{CoRR}, abs/1906.01749.

\bibitem[{Gao et~al.(2023)Gao, Dai, Pasupat, Chen, Chaganty, Fan, Zhao, Lao, Lee, Juan, and Guu}]{gao2023rarrresearchingrevisinglanguage}
Luyu Gao, Zhuyun Dai, Panupong Pasupat, Anthony Chen, Arun~Tejasvi Chaganty, Yicheng Fan, Vincent~Y. Zhao, Ni~Lao, Hongrae Lee, Da-Cheng Juan, and Kelvin Guu. 2023.
\newblock \href {https://arxiv.org/abs/2210.08726} {Rarr: Researching and revising what language models say, using language models}.
\newblock \emph{Preprint}, arXiv:2210.08726.

\bibitem[{Gekhman et~al.(2024)Gekhman, Yona, Aharoni, Eyal, Feder, Reichart, and Herzig}]{gekhman2024doesfinetuningllmsnew}
Zorik Gekhman, Gal Yona, Roee Aharoni, Matan Eyal, Amir Feder, Roi Reichart, and Jonathan Herzig. 2024.
\newblock \href {https://arxiv.org/abs/2405.05904} {Does fine-tuning llms on new knowledge encourage hallucinations?}
\newblock \emph{Preprint}, arXiv:2405.05904.

\bibitem[{Gliwa et~al.(2019)Gliwa, Mochol, Biesek, and Wawer}]{DBLP:journals/corr/abs-1911-12237}
Bogdan Gliwa, Iwona Mochol, Maciej Biesek, and Aleksander Wawer. 2019.
\newblock \href {https://arxiv.org/abs/1911.12237} {Samsum corpus: {A} human-annotated dialogue dataset for abstractive summarization}.
\newblock \emph{CoRR}, abs/1911.12237.

\bibitem[{Guerreiro et~al.(2023)Guerreiro, Voita, and Martins}]{guerreiro2023lookingneedlehaystackcomprehensive}
Nuno~M. Guerreiro, Elena Voita, and André F.~T. Martins. 2023.
\newblock \href {https://arxiv.org/abs/2208.05309} {Looking for a needle in a haystack: A comprehensive study of hallucinations in neural machine translation}.
\newblock \emph{Preprint}, arXiv:2208.05309.

\bibitem[{Guu et~al.(2020)Guu, Lee, Tung, Pasupat, and Chang}]{guu2020realmretrievalaugmentedlanguagemodel}
Kelvin Guu, Kenton Lee, Zora Tung, Panupong Pasupat, and Ming-Wei Chang. 2020.
\newblock \href {https://arxiv.org/abs/2002.08909} {Realm: Retrieval-augmented language model pre-training}.
\newblock \emph{Preprint}, arXiv:2002.08909.

\bibitem[{He et~al.(2023)He, Gao, and Chen}]{he2023debertav3improvingdebertausing}
Pengcheng He, Jianfeng Gao, and Weizhu Chen. 2023.
\newblock \href {https://arxiv.org/abs/2111.09543} {Debertav3: Improving deberta using electra-style pre-training with gradient-disentangled embedding sharing}.
\newblock \emph{Preprint}, arXiv:2111.09543.

\bibitem[{Islam et~al.(2023)Islam, Kannappan, Kiela, Qian, Scherrer, and Vidgen}]{islam2023financebenchnewbenchmarkfinancial}
Pranab Islam, Anand Kannappan, Douwe Kiela, Rebecca Qian, Nino Scherrer, and Bertie Vidgen. 2023.
\newblock \href {https://arxiv.org/abs/2311.11944} {Financebench: A new benchmark for financial question answering}.
\newblock \emph{Preprint}, arXiv:2311.11944.

\bibitem[{Izacard et~al.(2024)Izacard, Lewis, Lomeli, Hosseini, Petroni, Schick, Dwivedi-Yu, Joulin, Riedel, and Grave}]{10.5555/3648699.3648950}
Gautier Izacard, Patrick Lewis, Maria Lomeli, Lucas Hosseini, Fabio Petroni, Timo Schick, Jane Dwivedi-Yu, Armand Joulin, Sebastian Riedel, and Edouard Grave. 2024.
\newblock Atlas: few-shot learning with retrieval augmented language models.
\newblock \emph{J. Mach. Learn. Res.}, 24(1).

\bibitem[{Jin et~al.(2019)Jin, Dhingra, Liu, Cohen, and Lu}]{jin2019pubmedqadatasetbiomedicalresearch}
Qiao Jin, Bhuwan Dhingra, Zhengping Liu, William~W. Cohen, and Xinghua Lu. 2019.
\newblock \href {https://arxiv.org/abs/1909.06146} {Pubmedqa: A dataset for biomedical research question answering}.
\newblock \emph{Preprint}, arXiv:1909.06146.

\bibitem[{Joshi et~al.(2017)Joshi, Choi, Weld, and Zettlemoyer}]{joshi2017triviaqa}
Mandar Joshi, Eunsol Choi, Daniel~S Weld, and Luke Zettlemoyer. 2017.
\newblock Triviaqa: A large scale distantly supervised challenge dataset for reading comprehension.
\newblock \emph{arXiv preprint arXiv:1705.03551}.

\bibitem[{Kadavath et~al.(2022)Kadavath, Conerly, Askell, Henighan, Drain, Perez, Schiefer, Hatfield-Dodds, DasSarma, Tran-Johnson, Johnston, El-Showk, Jones, Elhage, Hume, Chen, Bai, Bowman, Fort, Ganguli, Hernandez, Jacobson, Kernion, Kravec, Lovitt, Ndousse, Olsson, Ringer, Amodei, Brown, Clark, Joseph, Mann, McCandlish, Olah, and Kaplan}]{kadavath2022languagemodelsmostlyknow}
Saurav Kadavath, Tom Conerly, Amanda Askell, Tom Henighan, Dawn Drain, Ethan Perez, Nicholas Schiefer, Zac Hatfield-Dodds, Nova DasSarma, Eli Tran-Johnson, Scott Johnston, Sheer El-Showk, Andy Jones, Nelson Elhage, Tristan Hume, Anna Chen, Yuntao Bai, Sam Bowman, Stanislav Fort, Deep Ganguli, Danny Hernandez, Josh Jacobson, Jackson Kernion, Shauna Kravec, Liane Lovitt, Kamal Ndousse, Catherine Olsson, Sam Ringer, Dario Amodei, Tom Brown, Jack Clark, Nicholas Joseph, Ben Mann, Sam McCandlish, Chris Olah, and Jared Kaplan. 2022.
\newblock \href {https://arxiv.org/abs/2207.05221} {Language models (mostly) know what they know}.
\newblock \emph{Preprint}, arXiv:2207.05221.

\bibitem[{Kembhavi et~al.(2017)Kembhavi, Seo, Schwenk, Choi, Farhadi, and Hajishirzi}]{Kembhavi2017AreYS}
Aniruddha Kembhavi, Minjoon Seo, Dustin Schwenk, Jonghyun Choi, Ali Farhadi, and Hannaneh Hajishirzi. 2017.
\newblock \href {https://api.semanticscholar.org/CorpusID:1310550} {Are you smarter than a sixth grader? textbook question answering for multimodal machine comprehension}.
\newblock \emph{2017 IEEE Conference on Computer Vision and Pattern Recognition (CVPR)}, pages 5376--5384.

\bibitem[{Kim et~al.(2024)Kim, Shin, Cho, Jang, Longpre, Lee, Yun, Shin, Kim, Thorne, and Seo}]{kim2024prometheusinducingfinegrainedevaluation}
Seungone Kim, Jamin Shin, Yejin Cho, Joel Jang, Shayne Longpre, Hwaran Lee, Sangdoo Yun, Seongjin Shin, Sungdong Kim, James Thorne, and Minjoon Seo. 2024.
\newblock \href {https://arxiv.org/abs/2310.08491} {Prometheus: Inducing fine-grained evaluation capability in language models}.
\newblock \emph{Preprint}, arXiv:2310.08491.

\bibitem[{Kornilova and Eidelman(2019)}]{DBLP:journals/corr/abs-1910-00523}
Anastassia Kornilova and Vlad Eidelman. 2019.
\newblock \href {https://arxiv.org/abs/1910.00523} {Billsum: {A} corpus for automatic summarization of {US} legislation}.
\newblock \emph{CoRR}, abs/1910.00523.

\bibitem[{Lai et~al.(2017)Lai, Xie, Liu, Yang, and Hovy}]{lai2017race}
Guokun Lai, Qizhe Xie, Hanxiao Liu, Yiming Yang, and Eduard Hovy. 2017.
\newblock Race: Large-scale reading comprehension dataset from examinations.
\newblock \emph{arXiv preprint arXiv:1704.04683}.

\bibitem[{Lee et~al.(2024)Lee, Phatale, Mansoor, Lu, Mesnard, Ferret, Bishop, Hall, Carbune, and Rastogi}]{lee2024rlaif}
Harrison Lee, Samrat Phatale, Hassan Mansoor, Kellie~Ren Lu, Thomas Mesnard, Johan Ferret, Colton Bishop, Ethan Hall, Victor Carbune, and Abhinav Rastogi. 2024.
\newblock \href {https://openreview.net/forum?id=AAxIs3D2ZZ} {{RLAIF}: Scaling reinforcement learning from human feedback with {AI} feedback}.

\bibitem[{Lewis et~al.(2020)Lewis, Perez, Piktus, Petroni, Karpukhin, Goyal, K\"{u}ttler, Lewis, Yih, Rockt\"{a}schel, Riedel, and Kiela}]{10.5555/3495724.3496517}
Patrick Lewis, Ethan Perez, Aleksandra Piktus, Fabio Petroni, Vladimir Karpukhin, Naman Goyal, Heinrich K\"{u}ttler, Mike Lewis, Wen-tau Yih, Tim Rockt\"{a}schel, Sebastian Riedel, and Douwe Kiela. 2020.
\newblock Retrieval-augmented generation for knowledge-intensive nlp tasks.
\newblock In \emph{Proceedings of the 34th International Conference on Neural Information Processing Systems}, NIPS '20, Red Hook, NY, USA. Curran Associates Inc.

\bibitem[{Li et~al.(2023)Li, Cheng, Zhao, Nie, and Wen}]{li-etal-2023-halueval}
Junyi Li, Xiaoxue Cheng, Xin Zhao, Jian-Yun Nie, and Ji-Rong Wen. 2023.
\newblock \href {https://doi.org/10.18653/v1/2023.emnlp-main.397} {{H}alu{E}val: A large-scale hallucination evaluation benchmark for large language models}.
\newblock In \emph{Proceedings of the 2023 Conference on Empirical Methods in Natural Language Processing}, pages 6449--6464, Singapore. Association for Computational Linguistics.

\bibitem[{Lin et~al.(2024)Lin, Gao, Oguz, Xiong, Lin, tau Yih, and Chen}]{lin2024flamefactualityawarealignmentlarge}
Sheng-Chieh Lin, Luyu Gao, Barlas Oguz, Wenhan Xiong, Jimmy Lin, Wen tau Yih, and Xilun Chen. 2024.
\newblock \href {https://arxiv.org/abs/2405.01525} {Flame: Factuality-aware alignment for large language models}.
\newblock \emph{Preprint}, arXiv:2405.01525.

\bibitem[{Lin et~al.(2022)Lin, Hilton, and Evans}]{lin-etal-2022-truthfulqa}
Stephanie Lin, Jacob Hilton, and Owain Evans. 2022.
\newblock \href {https://doi.org/10.18653/v1/2022.acl-long.229} {{T}ruthful{QA}: Measuring how models mimic human falsehoods}.
\newblock In \emph{Proceedings of the 60th Annual Meeting of the Association for Computational Linguistics (Volume 1: Long Papers)}, pages 3214--3252, Dublin, Ireland. Association for Computational Linguistics.

\bibitem[{Luo et~al.(2024)Luo, Shen, Li, Peng, Xuan, Wang, and Yang}]{luo2024halludiallargescalebenchmarkautomatic}
Wen Luo, Tianshu Shen, Wei Li, Guangyue Peng, Richeng Xuan, Houfeng Wang, and Xi~Yang. 2024.
\newblock \href {https://arxiv.org/abs/2406.07070} {Halludial: A large-scale benchmark for automatic dialogue-level hallucination evaluation}.
\newblock \emph{Preprint}, arXiv:2406.07070.

\bibitem[{Magesh et~al.(2024)Magesh, Surani, Dahl, Suzgun, Manning, and Ho}]{magesh2024hallucinationfreeassessingreliabilityleading}
Varun Magesh, Faiz Surani, Matthew Dahl, Mirac Suzgun, Christopher~D. Manning, and Daniel~E. Ho. 2024.
\newblock \href {https://arxiv.org/abs/2405.20362} {Hallucination-free? assessing the reliability of leading ai legal research tools}.
\newblock \emph{Preprint}, arXiv:2405.20362.

\bibitem[{Manakul et~al.(2023)Manakul, Liusie, and Gales}]{manakul2023selfcheckgptzeroresourceblackboxhallucination}
Potsawee Manakul, Adian Liusie, and Mark J.~F. Gales. 2023.
\newblock \href {https://arxiv.org/abs/2303.08896} {Selfcheckgpt: Zero-resource black-box hallucination detection for generative large language models}.
\newblock \emph{Preprint}, arXiv:2303.08896.

\bibitem[{McKenna et~al.(2023)McKenna, Li, Cheng, Hosseini, Johnson, and Steedman}]{mckenna2023sources}
Nick McKenna, Tianyi Li, Liang Cheng, Mohammad~Javad Hosseini, Mark Johnson, and Mark Steedman. 2023.
\newblock \href {https://openreview.net/forum?id=rJhk7Fpnvh} {Sources of hallucination by large language models on inference tasks}.
\newblock In \emph{The 2023 Conference on Empirical Methods in Natural Language Processing}.

\bibitem[{Mishra et~al.(2024)Mishra, Asai, Balachandran, Wang, Neubig, Tsvetkov, and Hajishirzi}]{mishra2024finegrainedhallucinationdetectionediting}
Abhika Mishra, Akari Asai, Vidhisha Balachandran, Yizhong Wang, Graham Neubig, Yulia Tsvetkov, and Hannaneh Hajishirzi. 2024.
\newblock \href {https://arxiv.org/abs/2401.06855} {Fine-grained hallucination detection and editing for language models}.
\newblock \emph{Preprint}, arXiv:2401.06855.

\bibitem[{M{\"o}ller et~al.(2020)M{\"o}ller, Reina, Jayakumar, and Pietsch}]{moller-etal-2020-covid}
Timo M{\"o}ller, Anthony Reina, Raghavan Jayakumar, and Malte Pietsch. 2020.
\newblock \href {https://aclanthology.org/2020.nlpcovid19-acl.18} {{COVID-QA}: A question answering dataset for {COVID}-19}.
\newblock In \emph{Proceedings of the 1st Workshop on {NLP} for {COVID-19} at {ACL} 2020}, Online. Association for Computational Linguistics.

\bibitem[{Nie et~al.(2020)Nie, Williams, Dinan, Bansal, Weston, and Kiela}]{nie-etal-2020-adversarial}
Yixin Nie, Adina Williams, Emily Dinan, Mohit Bansal, Jason Weston, and Douwe Kiela. 2020.
\newblock \href {https://doi.org/10.18653/v1/2020.acl-main.441} {Adversarial {NLI}: A new benchmark for natural language understanding}.
\newblock In \emph{Proceedings of the 58th Annual Meeting of the Association for Computational Linguistics}, pages 4885--4901, Online. Association for Computational Linguistics.

\bibitem[{Petroni et~al.(2021)Petroni, Piktus, Fan, Lewis, Yazdani, De~Cao, Thorne, Jernite, Karpukhin, Maillard, Plachouras, Rockt{\"a}schel, and Riedel}]{petroni-etal-2021-kilt}
Fabio Petroni, Aleksandra Piktus, Angela Fan, Patrick Lewis, Majid Yazdani, Nicola De~Cao, James Thorne, Yacine Jernite, Vladimir Karpukhin, Jean Maillard, Vassilis Plachouras, Tim Rockt{\"a}schel, and Sebastian Riedel. 2021.
\newblock \href {https://doi.org/10.18653/v1/2021.naacl-main.200} {{KILT}: a benchmark for knowledge intensive language tasks}.
\newblock In \emph{Proceedings of the 2021 Conference of the North American Chapter of the Association for Computational Linguistics: Human Language Technologies}, pages 2523--2544, Online. Association for Computational Linguistics.

\bibitem[{Ravi et~al.(2024)Ravi, Mielczarek, Kannappan, Kiela, and Qian}]{ravi2024lynxopensourcehallucination}
Selvan~Sunitha Ravi, Bartosz Mielczarek, Anand Kannappan, Douwe Kiela, and Rebecca Qian. 2024.
\newblock \href {https://arxiv.org/abs/2407.08488} {Lynx: An open source hallucination evaluation model}.
\newblock \emph{Preprint}, arXiv:2407.08488.

\bibitem[{Saad-Falcon et~al.(2024)Saad-Falcon, Khattab, Potts, and Zaharia}]{saadfalcon2024aresautomatedevaluationframework}
Jon Saad-Falcon, Omar Khattab, Christopher Potts, and Matei Zaharia. 2024.
\newblock \href {https://arxiv.org/abs/2311.09476} {Ares: An automated evaluation framework for retrieval-augmented generation systems}.
\newblock \emph{Preprint}, arXiv:2311.09476.

\bibitem[{Sander and Dietz(2021)}]{exam}
David~P Sander and Laura Dietz. 2021.
\newblock \href {https://ceur-ws.org/Vol-2950/paper-16.pdf} {Exam: How to evaluate retrieve-and-generate systems for users who do not (yet) know what they want}.

\bibitem[{Sharma et~al.(2019)Sharma, Li, and Wang}]{DBLP:journals/corr/abs-1906-03741}
Eva Sharma, Chen Li, and Lu~Wang. 2019.
\newblock \href {https://arxiv.org/abs/1906.03741} {{BIGPATENT:} {A} large-scale dataset for abstractive and coherent summarization}.
\newblock \emph{CoRR}, abs/1906.03741.

\bibitem[{Shuster et~al.(2021)Shuster, Poff, Chen, Kiela, and Weston}]{shuster-etal-2021-retrieval-augmentation}
Kurt Shuster, Spencer Poff, Moya Chen, Douwe Kiela, and Jason Weston. 2021.
\newblock \href {https://doi.org/10.18653/v1/2021.findings-emnlp.320} {Retrieval augmentation reduces hallucination in conversation}.
\newblock In \emph{Findings of the Association for Computational Linguistics: EMNLP 2021}, pages 3784--3803, Punta Cana, Dominican Republic. Association for Computational Linguistics.

\bibitem[{Tang et~al.(2024{\natexlab{a}})Tang, Laban, and Durrett}]{tang2024minicheck}
Liyan Tang, Philippe Laban, and Greg Durrett. 2024{\natexlab{a}}.
\newblock Minicheck: Efficient fact-checking of llms on grounding documents.
\newblock \emph{arXiv preprint arXiv:2404.10774}.

\bibitem[{Tang et~al.(2024{\natexlab{b}})Tang, Laban, and Durrett}]{tang2024minicheckefficientfactcheckingllms}
Liyan Tang, Philippe Laban, and Greg Durrett. 2024{\natexlab{b}}.
\newblock \href {https://arxiv.org/abs/2404.10774} {Minicheck: Efficient fact-checking of llms on grounding documents}.
\newblock \emph{Preprint}, arXiv:2404.10774.

\bibitem[{Tian et~al.(2023)Tian, Mitchell, Yao, Manning, and Finn}]{tian2023finetuninglanguagemodelsfactuality}
Katherine Tian, Eric Mitchell, Huaxiu Yao, Christopher~D. Manning, and Chelsea Finn. 2023.
\newblock \href {https://arxiv.org/abs/2311.08401} {Fine-tuning language models for factuality}.
\newblock \emph{Preprint}, arXiv:2311.08401.

\bibitem[{Trischler et~al.(2016)Trischler, Wang, Yuan, Harris, Sordoni, Bachman, and Suleman}]{trischler2016newsqa}
Adam Trischler, Tong Wang, Xingdi Yuan, Justin Harris, Alessandro Sordoni, Philip Bachman, and Kaheer Suleman. 2016.
\newblock Newsqa: A machine comprehension dataset.
\newblock \emph{arXiv preprint arXiv:1611.09830}.

\bibitem[{Vu et~al.(2023)Vu, Iyyer, Wang, Constant, Wei, Wei, Tar, Sung, Zhou, Le, and Luong}]{vu2023freshllmsrefreshinglargelanguage}
Tu~Vu, Mohit Iyyer, Xuezhi Wang, Noah Constant, Jerry Wei, Jason Wei, Chris Tar, Yun-Hsuan Sung, Denny Zhou, Quoc Le, and Thang Luong. 2023.
\newblock \href {https://arxiv.org/abs/2310.03214} {Freshllms: Refreshing large language models with search engine augmentation}.
\newblock \emph{Preprint}, arXiv:2310.03214.

\bibitem[{Vu et~al.(2024)Vu, Krishna, Alzubi, Tar, Faruqui, and Sung}]{vu2024foundationalautoraterstaminglarge}
Tu~Vu, Kalpesh Krishna, Salaheddin Alzubi, Chris Tar, Manaal Faruqui, and Yun-Hsuan Sung. 2024.
\newblock \href {https://arxiv.org/abs/2407.10817} {Foundational autoraters: Taming large language models for better automatic evaluation}.
\newblock \emph{Preprint}, arXiv:2407.10817.

\bibitem[{Wang et~al.(2024)Wang, Xu, Zhou, Xiong, and Joty}]{wang2024direct}
Peifeng Wang, Austin Xu, Yilun Zhou, Caiming Xiong, and Shafiq Joty. 2024.
\newblock Direct judgement preference optimization.
\newblock \emph{arXiv preprint arXiv:2409.14664}.

\bibitem[{Wu et~al.(2023)Wu, Zhu, Xu, Shum, Niu, Zhong, Song, and Zhang}]{wu2023ragtruth}
Yuanhao Wu, Juno Zhu, Siliang Xu, Kashun Shum, Cheng Niu, Randy Zhong, Juntong Song, and Tong Zhang. 2023.
\newblock Ragtruth: A hallucination corpus for developing trustworthy retrieval-augmented language models.
\newblock \emph{arXiv preprint arXiv:2401.00396}.

\bibitem[{Xu et~al.(2024)Xu, Jain, and Kankanhalli}]{xu2024hallucinationinevitableinnatelimitation}
Ziwei Xu, Sanjay Jain, and Mohan Kankanhalli. 2024.
\newblock \href {https://arxiv.org/abs/2401.11817} {Hallucination is inevitable: An innate limitation of large language models}.
\newblock \emph{Preprint}, arXiv:2401.11817.

\bibitem[{Zheng et~al.(2023)Zheng, Chiang, Sheng, Zhuang, Wu, Zhuang, Lin, Li, Li, Xing, Zhang, Gonzalez, and Stoica}]{zheng2023judgingllmasajudgemtbenchchatbot}
Lianmin Zheng, Wei-Lin Chiang, Ying Sheng, Siyuan Zhuang, Zhanghao Wu, Yonghao Zhuang, Zi~Lin, Zhuohan Li, Dacheng Li, Eric~P. Xing, Hao Zhang, Joseph~E. Gonzalez, and Ion Stoica. 2023.
\newblock \href {https://arxiv.org/abs/2306.05685} {Judging llm-as-a-judge with mt-bench and chatbot arena}.
\newblock \emph{Preprint}, arXiv:2306.05685.

\bibitem[{Zhu et~al.(2023)Zhu, Wang, and Wang}]{zhu2023judgelmfinetunedlargelanguage}
Lianghui Zhu, Xinggang Wang, and Xinlong Wang. 2023.
\newblock \href {https://arxiv.org/abs/2310.17631} {Judgelm: Fine-tuned large language models are scalable judges}.
\newblock \emph{Preprint}, arXiv:2310.17631.

\end{thebibliography}

\clearpage

\appendix

\onecolumn

\newpage
\section{VERITAS DeBERTa Input Format}
\label{sec:appendix_deberta_input_format}

\begin{tcolorbox}[
  colback=gray!20,
  colframe=gray!60!black,
  title= DeBERTa Input Format For NLI,
  fonttitle=\bfseries\large
]

\begin{lstlisting}[style=jinja2]"""
# Context:
{{ document }}

# Claim:
assistant: {{ claim }}                                  
\end{lstlisting}
\end{tcolorbox}

\begin{tcolorbox}[
  colback=gray!20,
  colframe=gray!60!black,
  title= DeBERTa Input Format For QA,
  fonttitle=\bfseries\large
]

\begin{lstlisting}[style=jinja2]"""
# Context:
{{ document }}

# Claim:
user: {{ question }}
assistant: {{ answer }}                                   
\end{lstlisting}
\end{tcolorbox}

\begin{tcolorbox}[
  colback=gray!20,
  colframe=gray!60!black,
  title= DeBERTa Input Format For Dialogue,
  fonttitle=\bfseries\large
]

\begin{lstlisting}[style=jinja2]"""
# Context:
{{ document }}

# Claim:
{% for message in conversation %}
{{ message.role }}: {{ message.content }}
{% endfor %}                              
\end{lstlisting}
\end{tcolorbox}

\begin{tcolorbox}[
  colback=gray!20,
  colframe=gray!60!black,
  title= DeBERTa Input Format For Summary Verification,
  fonttitle=\bfseries\large
]

\begin{lstlisting}[style=jinja2]"""
# Context:
{{ document }}

# Summary:
assistant: {{ summary }}                            
\end{lstlisting}
\end{tcolorbox}

\section{VERITAS LLama Instruction Format}
\label{sec:appendix_llama_input_format}

\begin{tcolorbox}[
  colback=gray!20,
  colframe=gray!60!black,
  title= LLama Input Format For NLI,
  fonttitle=\bfseries\large
]

\begin{lstlisting}[style=jinja2]"""
You will classify whether the claim is supported by the given document or not.

Follow these steps:\n\n
1. Assess the claim against the document \n
2. Classify it is 0 (not supported) or 1 (supported)  \n
3. Provide Rationale: Explain your classification decision with a brief rationale.\n
4. Finally, output the results as a JSON object with the fields \"rationale\" and \"output\" where \"output\" contains the classification (0 or 1)

# Document:\n
{{ document }}\n

# Claim:\n
{{ claim }}\n

Now, please output the following as a JSON object:\n
{
"rationale": <verbal feedback> (str datatype),\n  
"output": <classification score (0 or 1)> (int datatype),\n
}           
\end{lstlisting}
\end{tcolorbox}

\begin{tcolorbox}[
  colback=gray!20,
  colframe=gray!60!black,
  title= LLama Input Format For QA,
  fonttitle=\bfseries\large
]

\begin{lstlisting}[style=jinja2]"""
You will classify whether the answer is supported by the given document or not.

Follow these steps:\n\n
1. Assess the answer against the document \n
2. Classify it is 0 (not supported) or 1 (supported)  \n
3. Provide Rationale: Explain your classification decision with a brief rationale.\n
4. Finally, output the results as a JSON object with the fields \"rationale\" and \"output\" where \"output\" contains the classification (0 or 1)

# Document:\n
{{ document }}\n

# Question:\n
{{ question }}\n

# Answer:\n
{{ answer }}\n

Now, please output the following as a JSON object:\n
{
"rationale": <verbal feedback> (str datatype),\n  
"output": <classification score (0 or 1)> (int datatype),\n
}           
\end{lstlisting}
\end{tcolorbox}

\begin{tcolorbox}[
  colback=gray!20,
  colframe=gray!60!black,
  title= LLama Input Format For Dialogue,
  fonttitle=\bfseries\large
]

\begin{lstlisting}[style=jinja2]"""
You will classify whether the last assistant response in the conversation is supported by the given document or not.

Follow these steps:\n\n
1. Assess the last assistant response in the conversation against the document \n
2. Classify it is 0 (not supported) or 1 (supported)  \n
3. Provide Rationale: Explain your classification decision with a brief rationale.\n
4. Finally, output the results as a JSON object with the fields \"rationale\" and \"output\" where \"output\" contains the classification (0 or 1)

# Document:\n
{{ document }}\n

# Conversation:\n
{% for message in conversation %}
{{ message.role }}: {{ message.content }}
{% endfor %}

Now, please output the following as a JSON object:\n
{
"rationale": <verbal feedback> (str datatype),\n  
"output": <classification score (0 or 1)> (int datatype),\n
}           
\end{lstlisting}
\end{tcolorbox}

\begin{tcolorbox}[
  colback=gray!20,
  colframe=gray!60!black,
  title= LLama Input Format For Summary Verification,
  fonttitle=\bfseries\large
]

\begin{lstlisting}[style=jinja2]"""
You will classify whether the given summary is supported by the given document or not.

Follow these steps:\n\n
1. Assess the summary against the document \n
2. Classify it is 0 (not supported) or 1 (supported)  \n
3. Provide Rationale: Explain your classification decision with a brief rationale.\n
4. Finally, output the results as a JSON object with the fields \"rationale\" and \"output\" where \"output\" contains the classification (0 or 1)

# Document:\n
{{ document }}\n

# Summary:\n
{{ summary }}\n

Now, please output the following as a JSON object:\n
{
"rationale": <verbal feedback> (str datatype),\n  
"output": <classification score (0 or 1)> (int datatype),\n
}           
\end{lstlisting}
\end{tcolorbox}
\clearpage

\enlargethispage{\baselineskip}

\end{document}